\documentclass[letterpaper, 10 pt, conference]{ieeeconf}  %
\usepackage{cite}
\usepackage{xcolor} 
\usepackage{booktabs}
\usepackage{graphicx}
\usepackage{amsfonts}
\usepackage{amsmath}
\usepackage{xspace}
\usepackage{bm}
\usepackage{verbatim}
\usepackage[ruled,commentsnumbered,linesnumbered,noend]{algorithm2e}
\usepackage{multirow}
\usepackage{array}
\usepackage{url}
\newcommand{\method}{Q-DEG\xspace}

\newif\ifqdshowchanges
\qdshowchangestrue

\IEEEoverridecommandlockouts                              %

\title{\LARGE \bf
Red-Teaming of Vision-Language-Action Models\\via Quality Diversity Environment Generation
}

\newif\ificraanonymous
\icraanonymousfalse   %

\ificraanonymous
  \author{Anonymous Authors}
\else
    \author{Shihan Zhao$^{1}$$^*$, Henry Chen$^{1}$, Ya-Chuan Hsu$^{1}$, Minjune Hwang$^{1}$, Siddharth Srikanth$^{12}$\\ Freddie Liang$^{1}$, Akanksha Saran$^{3}$, Aaquib Tabrez$^{4}$$^{\dagger}$, Stefanos Nikolaidis$^{1}$$^{\dagger}$, Daniel Seita$^{1}$$^{\dagger}$%
    \thanks{$^{1}$Thomas Lord Department of Computer Science, University of Southern California. $^{2}$Columbia University. $^{3}$Adobe Research $^{4}$Cornell University. $^{\dagger}$Equal advisement. $^*$Corresponding author. Inquiries should be emailed to  {\tt\small shihanzh@usc.edu}.}%
    }%
\fi

\begin{document}
\bstctlcite{IEEEexample:BSTcontrol}

\maketitle
\thispagestyle{empty}
\pagestyle{empty}

\begin{abstract}
A long-standing goal in robotics is to develop generalist robots that can perform tasks across a wide range of settings, flexibly specified with images and text. Vision-Language-Action (VLA) models offer a promising path toward this goal by training on large collections of robot demonstrations. However, while VLAs perform well in-distribution, they often fail under minor test-time shifts, such as changes in object positions, lighting, textures, and camera poses.
To address this, we propose \textbf{\underline{\smash{Q}}}uality Diversity for \textbf{\underline{\smash{D}}}iverse \textbf{\underline{\smash{E}}}nvironment \textbf{\underline{\smash{G}}}eneration (\method), 
a method for improving VLA robustness by generating adversarial environments and fine-tuning the VLA on its own successful rollouts.
\method uses Quality Diversity (QD) optimization to find a set of high-failure environments, while promoting diversity using policy embeddings extracted from the VLA. These environments expose vulnerabilities that can be mitigated through self-imitation fine-tuning, thereby improving the VLA's robustness.
In simulation, we conduct experiments across three representative VLA models, and our results indicate that \method improves robustness to out-of-distribution (OOD) adversarial environments compared to the base VLA and VLAs fine-tuned using data collected from two red-teaming baselines. Furthermore, we demonstrate that improving VLA robustness through \method facilitates sim-to-real transfer. Our anonymous project website is at: \url{https://qd-env.github.io/qdeg/}.
\end{abstract}

\section{Introduction}

Vision-Language-Action (VLA) models~\cite{kim2024openvla, black_0_nodate,pertsch2025fastefficientactiontokenization,intelligence2025pi05,shukor2025smolvla,lee2025molmoact, fang2026molmoact2actionreasoningmodels} have emerged as a promising mode for robot control. By leveraging Vision-Language Models (VLMs) pretrained on internet-scale data, VLAs have access to cross-domain visual and semantic knowledge, and generalize across multiple tasks and robot embodiments. However, despite strong in-distribution performance, VLAs frequently fail when test scenarios differ from training data in peripheral ways such as object positions, textures and colors, lighting, and camera poses. 

Prior work has characterized this vulnerability by analyzing VLA failure modes in OOD environments~\cite{gao2025taxonomy, zhou2025libero}, but these environments are hand-designed for benchmarking and not designed to elicit failures in a specific VLA and improve its robustness. Fine-tuning VLAs using reinforcement learning objectives can improve their robustness end-to-end~\cite{liu2025can, lu2025vla, chen2025conrft}, but these methods do not expose model vulnerabilities by identifying failure-inducing environments. Red-teaming methods probe policy-specific weaknesses through generative visual edits~\cite{majumdar2025predictiveredteaming} and video world model rollouts~\cite{geminiroboticsteam2026evaluating}, but they do not search over interpretable environment parameters in a physics simulator.

\begin{figure*}[t]
\centering
\scalebox{1}[0.95]{%
  \includegraphics[width=0.95\linewidth]{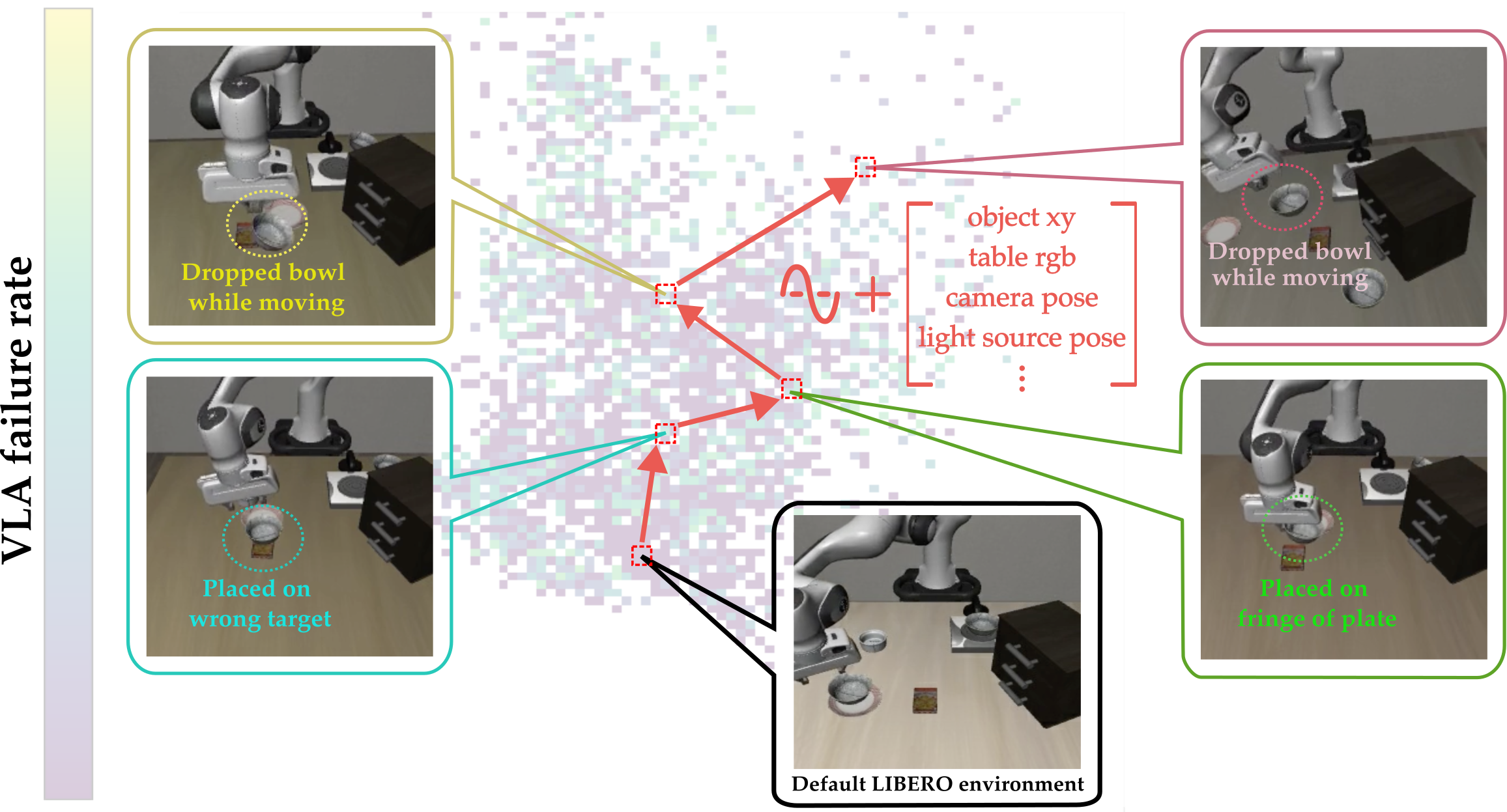}%
}
\vspace*{-10pt}
\caption{
  \method generates diverse failure-inducing environments by perturbing parameters such as object placement, background, and camera and light source poses. New perturbations are generated from previously discovered environments, which act as ``stepping stones'' that facilitate exploration.
}
\vspace*{-10pt}
\label{fig:pull}
\end{figure*}

Prior work established Quality Diversity (QD) optimization as a framework for testing robotic systems by discovering diverse scenarios~\cite{fontaine2021quality,fontaine2021importanceenvironment,nikolaidis2024algorithmic,samvelyan2024rainbow,srikanth2026redteaming}. We extend QD to VLA environmental red-teaming and propose \textbf{\underline{\smash{Q}}}uality Diversity for \textbf{\underline{D}}iverse \textbf{\underline{E}}nvironment \textbf{\underline{G}}eneration (\method). Q-DEG searches for environment parameterizations that cause a target VLA to fail frequently while inducing diverse initial policy embeddings, extracted from the VLA's pre-action-layer activations. The search prioritizes environments that still yield occasional successful rollouts and caches these trajectories for subsequent self-imitation fine-tuning to improve VLA robustness. \method organizes the generated environments into a heatmap (Figure~\ref{fig:pull}) that maps policy embeddings to success rates, facilitating analysis of model vulnerabilities. It uses episode-level task success as its performance feedback and policy embeddings for diversity, requiring no action log-probabilities and thus supporting both autoregressive and flow-based VLAs.

To summarize, our contributions are: (1) \method, a method that formulates VLA environmental red-teaming as a Quality Diversity search over environment parameterizations; (2) a self-imitation~\cite{oh2018self} procedure that improves VLA robustness by fine-tuning on the VLA's own successful rollouts saved during QD search; and (3) an evaluation of \method across three VLAs in LIBERO~\cite{liu2023libero}, showing advantage over baselines in environmental diversity, robustness improvement, and sim-to-real transfer.

\section{Related Works}

\paragraph{Vision-Language-Action (VLA) Models} 
VLAs aim to provide general-purpose robot control by mapping visual observations and language instructions directly to robot actions.  
Examples of such VLAs include OpenVLA~\cite{kim2024openvla}, the $\pi$ family~\cite{black_0_nodate,pertsch2025fastefficientactiontokenization,intelligence2025pi05}, SmolVLA~\cite{shukor2025smolvla}, and MolmoAct models~\cite{lee2025molmoact, fang2026molmoact2actionreasoningmodels}. 
These models are trained on broad robot datasets for language-conditioned manipulation. Techniques such as fine-tuning~\cite{kim2025fine} and steering~\cite{chen2026steerableVLA} further improve their success rate and adaptability on downstream tasks. 
However, making VLAs robust remains challenging because even large robot datasets~\cite{khazatsky2024droid,open_x_embodiment_rt_x_2024} cover only a limited subset of possible object poses, lighting conditions, camera viewpoints, textures, and scene layouts. Consequently, VLAs remain brittle~\cite{liu2025can, gao2025taxonomy, zhou2025libero} under minor environment shifts even when they do not alter the semantic task. To address this, \method searches for \emph{physical environment variations} that expose failures, then uses the VLA's successful rollouts in those environments to improve robustness.

\paragraph{Quality Diversity Optimization} Quality Diversity (QD) methods search for collections of high-quality solutions that differ along specified measures, often by discretizing the measure space into an archive of cells to be filled through iterative evolution~\cite{mouret2015illuminating}. The search efficiency may be improved with covariance matrix adaptation~\cite{fontaine2023covariance, zhao2025covariance}, gradient-based optimization~\cite{fontaine2021differentiable}, and in-context learning~\cite{lim2024largelanguagemodelsincontext}. Compared with pure optimization, QD’s diversity incentive makes it well suited to generating scenarios and inputs that elicit diverse responses from a policy. QD has been used to generate scenarios that probe collaborative strategies in multi-agent systems~\cite{fontaine2021importanceenvironment} and expose failures in robot-assisted work settings~\cite{fontaine2021quality,pmlr-v229-bhatt23a}. See~\cite{nikolaidis2024algorithmic} for a review of QD-based scenario generation and testing. These works use QD primarily to characterize a policy’s behavior and limitations with respect to user-specified diversity measures. In contrast, \method searches for diverse, challenging scenarios to generate training data for improving a VLA’s robustness, using the VLA’s policy embeddings as diversity measures. 
 
 Recently, QD has also been applied to instruction red-teaming for foundation models. Rainbow Teaming~\cite{samvelyan2024rainbow} generates prompts that elicit unsafe responses from an LLM, and Q-DIG~\cite{srikanth2026redteaming} generates instructions that describe the task accurately but still cause a VLA to fail. \method complements these works by extending QD to environmental red-teaming for VLAs, searching over environment parameters that induce VLA failures while keeping the instruction fixed.

\paragraph{Environment Design}
A complementary line of work develops simulated environments and benchmarks for evaluating language-conditioned robot policies.  
Benchmarks such as LIBERO~\cite{liu2023libero}, LIBERO-PRO~\cite{zhou2025libero}, and RoboCasa~\cite{nasiriany2024robocasa} provide valuable testbeds for measuring generalization across tasks, scenes, and physical reasoning challenges. However, these benchmarks are mainly designed for evaluation, with environment variations specified by benchmark designers rather than searched to target a VLA's failure modes, and the resulting failures are not directly converted into fine-tuning data.  %
Other works on unsupervised environment design automatically generate challenging tasks or curricula for reinforcement learning agents~\cite{jiang2021prioritized,parker2022evolving} but do not target modern VLA policies. %
Complementary works study geometric red-teaming for robot manipulation~\cite{goel2025redteaming}, generative simulation for robot learning~\cite{katara2024gen2sim}, and world simulators for policy evaluation~\cite{geminiroboticsteam2026evaluating}. 

Within this landscape, \method searches over interpretable simulator parameters, using QD to promote diversity among the VLA policy embeddings induced by the resulting environments.

\begin{figure*}[t]
\centering
\scalebox{1}[0.95]{%
  \includegraphics[width=0.95\linewidth]{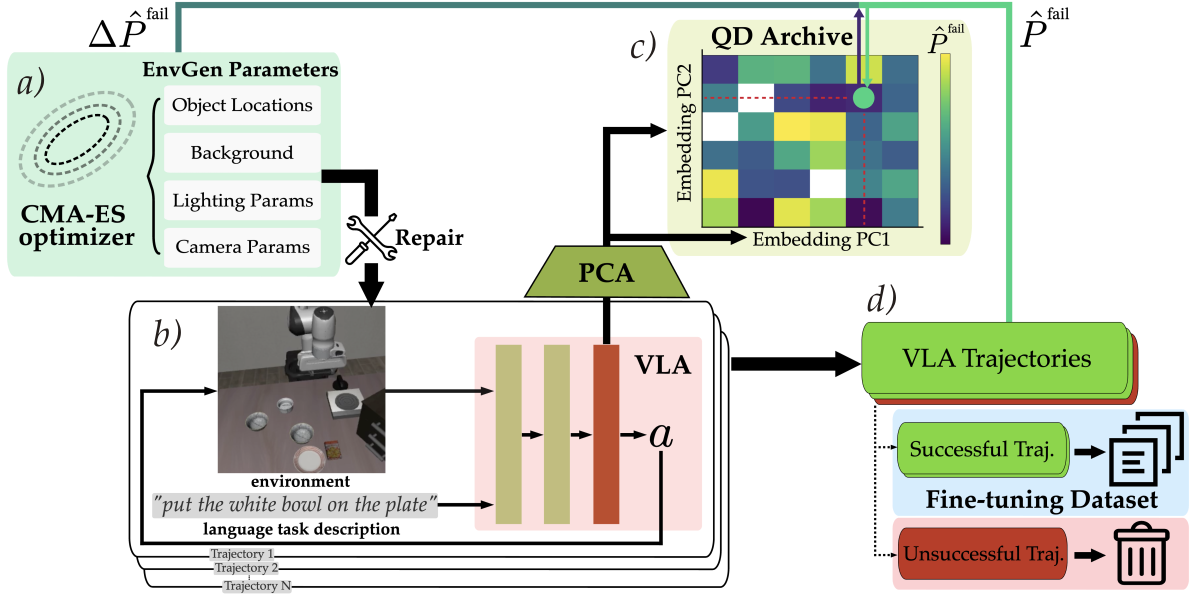}
}
\vspace*{-10pt}
\caption{
  \textbf{Overview of \method}. \textit{a)} CMA-ES~\cite{hansen2023cmaevolutionstrategy} proposes environment parameters such as object locations, background, lighting, and camera pose; invalid proposals are repaired to restore task-relevant spatial relations. \textit{b)} The VLA is rolled out in each proposed environment to obtain feasibility-aware failure rate $\hat{P}^\text{fail}$ and policy embeddings. \textit{c)} \method adds each environment to the QD archive based on its policy embeddings, and rewards CMA-ES with its failure rate \textit{improvement} $\Delta\hat{P}^\text{fail}$ over predecessors with similar embeddings. \textit{d)} Successful trajectories from archived environments are collected into a fine-tuning dataset for improving the VLA’s robustness.
}
\vspace*{-10pt}
\label{fig:method}
\end{figure*}

\section{Problem Formulation and Assumptions}
\label{sec:problem_formulation}

We assume the following as part of our problem setting. (1) A visual-semantic robot task $\mathcal{E}_{\bm{c}}(\theta)$, defined by a language instruction $\bm{c}$ and parameterized by interpretable environment parameters (e.g., object and camera poses) $\theta \in \Theta$. (2) A task success indicator $\mathbf{1}^{\text{succ}}(\zeta) \in \{0,1\}$, where $\zeta$ denotes a trajectory. (3) A base Vision-Language-Action (VLA) model $\pi(\bm{a} \mid \bm{o}, \bm{c})$ that maps an observation $\bm{o}$ and instruction $\bm{c}$ to a robot action $\bm{a}$ and is capable of succeeding on $\mathcal{E}_{\bm{c}}(\theta_0)$, where $\theta_0 \in \Theta$ is the default task parameterization. Under this setting, the failure rate of the base VLA is
\begin{equation}
P^\text{fail}(\theta) = \mathbb{E}_{\zeta \sim \pi,\,\mathcal{E}_{\bm{c}}(\theta)}
\left[1-\mathbf{1}^\text{succ}(\zeta)\right].
\end{equation}
We address two problems: (1) Under a fixed $\bm{c}$, generate a collection of diverse environments parameterized by $\{\theta_i \in \Theta\}$ in which the base VLA achieves high failure rates $P^\text{fail}(\theta_i)$. We exclude maximally adversarial environments that yield a $100\%$ failure rate to prioritize \textit{feasibly adversarial} environments near the boundary of the base VLA's capabilities. (2) Improve the base VLA's robustness to environment variations within $\Theta$.

\section{Method: \method}
\label{sec:method}

\method improves a base VLA by using a parameterized simulator to propose and repair task-consistent environment configurations (Sec.~\ref{ssec:envgen}), rolling out the VLA to estimate feasibility-aware failure rates and extract policy embeddings (Sec.~\ref{ssec:feedback_metrics}), archiving diverse adversarial environments through QD search (Sec.~\ref{ssec:qd_optimization}), and fine-tuning the VLA on its own successful rollouts from those environments (Sec.~\ref{ssec:fine_tuning}). See Figure~\ref{fig:method} for an overview.

\subsection{Environment Generation}
\label{ssec:envgen}
At the start of each iteration, \method generates a batch of $B$ candidate environments parameterized by $\{\theta_i\}_{i=1}^B$. Following common axes of variation studied for policy robustness~\cite{chen2026craftvideodiffusionbimanual}, we perturb object placements ($xy$ coordinates of each object), camera pose ($xyz$ coordinates and axis-angle orientation), light-source position ($xyz$ coordinates), table-background RGB, and reflective specular RGB, yielding a 25-dimensional continuous search space. To prevent these perturbations from violating the original task semantics, we repair each candidate $\theta_i$ using a mixed-integer quadratic program (MIQP)~\cite{nemhauser1988integer} that enforces semantic constraints while minimizing edit distance $\|\theta_i^\text{repaired} - \theta_i\|_2^2$. For example, if the task is ``pick up bowl next to plate,'' but a perturbed candidate places the two objects far apart, MIQP will minimally modify the objects' coordinates such that the distance between them is sufficiently small to be considered as ``close.''
\method is not tied to the parameterization considered in this work: different parameters may be perturbed depending on the robustness axes of interest, and MIQP repair is unnecessary when the perturbed parameters do not affect task semantics (e.g., when only background color is perturbed).

\subsection{Feedback Metrics}
\label{ssec:feedback_metrics}
On each of the $B$ proposed environments, the base VLA $\pi$ is rolled out $k$ times to obtain two feedback metrics.

\noindent\textbf{Failure Rate:} To evaluate an environment's adversarial \textit{quality} with respect to the VLA $\pi$, $\pi$'s empirical failure rate $\hat{P}^\text{fail}$ on the environment is estimated over trajectories $\{\zeta^j\}_{j=1}^{k}$. \method prioritizes environments that are feasible enough to yield successful rollouts for later self-imitation, so we modify $\hat{P}^\text{fail}$ with a condition that hard-sets it to $0$ if all rollouts fail, as defined in Eq.~\ref{eq:pfail}.
\vspace{-5pt}
\begin{equation}
\label{eq:pfail}
\hat{P}^\text{fail}(\theta)=
\begin{cases}
0 & \text{if } \bigwedge_{j=1}^{k} \neg\mathbf{1}^{\text{succ}}(\zeta^j) \\
1-\sum_{j=1}^k \mathbf{1}^{\text{succ}}(\zeta^j) / k & \text{otherwise.}
\end{cases}
\end{equation}

\noindent\textbf{Policy Embeddings:} For evaluating an environment's \textit{diversity}, we use $\pi$'s pre-action layer activations at the start of each rollout, which we refer to as \textit{policy embeddings}. Intuitively, policy embeddings represent the VLA's ``first impression'' on each environment, and our goal is to diversify policy embeddings so that our adversarial environments induce diverse experience on the VLA. Our choice of pre-action-layer activations as policy embeddings is motivated by prior work on foundation model anomaly detection~\cite{gu2506safe, ming2022delving}, which suggests that intermediate activations encode rich information about the model's internal processing and exhibit behaviorally meaningful local structure. We follow SAFE~\cite{gu2506safe} and use activations from the final hidden layer before the VLA's action decoder as policy embeddings. For each environment, we mean-pool these activations across $k$ rollouts to obtain an embedding vector $\psi$. Since raw embeddings are high-dimensional and QD algorithms often suffer from the curse of dimensionality~\cite{paolo2020unsupervised, hedayatian2025soft}, we project raw policy embeddings into a low-dimensional space using linear Principal Component Analysis (PCA) before QD search. We denote the projected embedding with $\mathbf{A}\psi + \mathbf{b}$, with $\mathbf{A}$ being the principal component projection matrix and $\mathbf{b}$ being the centering offset. See App.~\ref{app:polemb} for more details.

\subsection{Quality Diversity (QD) Optimization}
\label{ssec:qd_optimization}

\method populates an archive $\mathcal{A}$ with environment parameterizations $\{\theta_i\}$ that induce high VLA failure rates $\{\hat{P}^{\mathrm{fail}}(\theta_i)\}$ and diverse policy embeddings $\{\psi_i\}$ well dispersed throughout the VLA policy embedding space. Following QD conventions~\cite{mouret2015illuminating}, we define $\mathcal{A}$ as a grid of cells obtained by discretizing the embedding space at a specified resolution. Each cell stores environments whose policy embeddings fall within its embedding space region. The sum of the highest failure rate found within each occupied cell captures the archive's adversarial coverage across distinct policy embeddings, without allowing similar environments to inflate the objective more than once. \method therefore maximizes
\begin{equation}
\label{eq:qd_score}
\sum_{\text{cell} \in \mathcal{A}} \max_{i: \psi_i \in \text{cell}} \hat{P}^\text{fail}(\theta_i)
\end{equation}
We use Covariance Matrix Adaptation Evolution Strategy (CMA-ES)~\cite{hansen2023cmaevolutionstrategy}, a black-box optimizer effective in low-dimensional continuous search spaces with rugged objective landscapes, to search for candidate environments that improve Eq.~\ref{eq:qd_score}. CMA-ES maintains a multivariate Gaussian distribution over environment parameters and samples a batch of environments $\{\theta_i\}_{i=1}^B$ at each iteration, as described in Sec.~\ref{ssec:envgen}. It then adapts this distribution using scalar feedback from the sampled candidates. A natural feedback signal is each candidate's failure-rate improvement over the highest-failure incumbent in its assigned cell, corresponding to its marginal contribution to Eq.~\ref{eq:qd_score}. However, because $\hat{P}^{\text{fail}}$ is estimated from few rollouts, it takes on a small number of discrete values. Improvement over the incumbent therefore produces sparse and frequently tied feedback, providing little signal for CMA-ES adaptation.

To provide denser feedback, we follow the QD algorithm CMA-MAE~\cite{fontaine2023covariance,zhao2025covariance} and associate each cell with a \textit{threshold} $t_{\text{cell}}$, initialized to $0$. After evaluating $\hat{P}^{\text{fail}}(\theta_i)$ and $\psi_i$, we assign $\theta_i$ to $\text{cell}_i \gets \mathcal{T}(\mathbf{A}\psi_i + \mathbf{b})$, where $\mathcal{T}$ maps PCA-projected policy embeddings to cell indices. We then return to CMA-ES the threshold-relative \textit{improvement score} $\Delta \hat{P}^{\text{fail}}(\theta_i) = \hat{P}^{\text{fail}}(\theta_i) - t_{\text{cell}_i}$. After receiving a candidate, the cell's threshold is updated according to $t_{\text{cell}_i} \gets (1-\alpha)t_{\text{cell}_i} + \alpha\max\bigl(\hat{P}^{\text{fail}}(\theta_i), t_{\text{cell}_i}\bigr)$. The threshold rises gradually as more high-failure candidates are observed, rather than immediately jumping to the highest observed failure rate. The threshold can therefore take on intermediate values that reflect how consistently high-failure candidates have been found within the cell. Consequently, a candidate receives a high improvement score when it induces a high failure rate or maps to a cell with few prior high-failure candidates. Using these threshold-relative \textit{improvement scores} as feedback, CMA-ES adapts its sampling distribution toward environments that are more adversarial or induce novel policy embeddings.

If a candidate undergoes MIQP repair, we subtract its edit distance from its CMA-ES feedback to discourage the generation of invalid environments. When CMA-ES stagnates (i.e. no sampled environment has positive improvement score), we sample an occupied cell from the archive and restart CMA-ES by centering its distribution on the cell's highest-failure environment parameterization. This resumes QD search from a ``stepping stone'' environment known to be the most adversarial within an embedding space region.

\subsection{Self-Imitation Fine-Tuning}
\label{ssec:fine_tuning}
\method also improves the robustness of the base VLA $\pi$ to failure modes uncovered during QD search. It does so by fine-tuning $\pi$ to imitate its own successful trajectories on generated adversarial environments. Specifically, for each environment in $\mathcal{A}$ satisfying $0 < \hat{P}^{\mathrm{fail}} < 1$, we retain the shortest successful trajectory among its $k$ rollouts to construct the fine-tuning dataset $\mathcal{D}$. Environments with $\hat{P}^{\mathrm{fail}}=0$ are not adversarial, whereas those with $\hat{P}^{\mathrm{fail}}=1$ provide no successful trajectory to imitate. We use only the shortest successful trajectory per environment while excluding longer trajectories, as multiple different demonstrations on the same environment can lead to action divergence, which undermines imitation learning~\cite{belkhale2023data}. To mitigate forgetting, we add to $\mathcal{D}$ $10\%$ of $\pi$'s original training data for the default tasks~\cite{liu2026pretrained}. We then fine-tune $\pi$ on $\mathcal{D}$ using standard behavior cloning. Notably, the environmental diversity promoted by \method translates into trajectory diversity within $\mathcal{D}$, helping prevent behavior collapse during self-imitation~\cite{yuan2023scaling}. Alg.~\ref{alg:pseudocode} summarizes our method.

\begin{algorithm}[h]
\SetAlgoLined
\caption{\method}
\label{alg:pseudocode}
\SetKwInOut{Input}{input}
\SetKwInOut{Result}{result}
\DontPrintSemicolon

\Input{QD iterations $N$, measure space dimension $M$, CMA-ES batch size $B$, base VLA $\pi$, rollouts per environment $k$, initial environment parameters $\theta_0$, CMA-ES initial step size $\sigma_0$, CMA-MAE learning rate $\alpha$.}
\Result{Archive of adversarial environments $\mathcal{A}$; fine-tuning dataset $\mathcal{D}$.}

Initialize CMA-ES with distribution $\mathcal{N}(\theta_0, \sigma_0^2 I)$.

Initialize CMA-MAE thresholds $\{t_\text{cell} \gets 0\}_{\text{cell} \in \mathcal{A}}$.

Collect sample policy embeddings $\{\psi_i\}$ by rolling out $\pi$ for one
step in random environments.\;
$(\mathbf A,\mathbf b)\gets\operatorname{PCA}(\{\psi\},M)$\;

\For{$n\gets1$ \KwTo $N$}{
    $\{\theta_i\}_{i=1}^{B}\gets
    \operatorname{CMA-ES-Ask}(B)$\;
    
    $\{\theta_i^\text{repaired}\}_{i=1}^{B} \gets \operatorname{Repair}(\{\theta_i\}_{i=1}^{B})$\;

    \For{$i\gets1$ \KwTo $B$}{
        Obtain failure rate $\hat P_i^{\mathrm{fail}}$, policy embeddings
        $\psi_i$, and trajectories $\{\zeta^j_i\}_{j=1}^k$ by rolling out
        $\pi$ on environment $\theta_i^\text{repaired}$ $k$ times.\;

        $\text{cell}_i\gets\mathcal{T}(\mathbf{A}\psi_i+\mathbf{b})$\;
        $\mathcal{A}[\text{cell}_i]\gets\theta_i^\text{repaired}$\;
        $\Delta\hat{P}^\text{fail}_i\gets\hat P_i^{\mathrm{fail}}-t_{\text{cell}_i} - \|\theta_i^\text{repaired} - \theta_i\|_2^2$\;
        $t_{\text{cell}_i} \gets (1 - \alpha)t_{\text{cell}_i} + \alpha\max(\hat P_i^{\mathrm{fail}}, t_{\text{cell}_i})$\;

        \If{$0 < \hat P_i^{\mathrm{fail}} < 1$}{
            Add the shortest successful $\zeta$ to $\mathcal{D}$\;
        }
    }

    $\operatorname{CMA-ES-Tell}
    (\{(\theta_i,\Delta\hat{P}^\text{fail}_i)\}_{i=1}^{B})$\;

    \If{$\operatorname{CMA-ES-Stagnate}()$}{
        Reset CMA-ES to $\mathcal{N}(\theta^*, \sigma_0^2 I)$ around the highest-failure env. $\theta^*$ from $\text{cell} \sim \mathcal{A}$\;
    }
}
\end{algorithm}

\section{Experiments}
\label{sec:experiments}
Our experiments are designed to answer the following questions: 1) \textbf{Red-teaming:} Can \method generate diverse adversarial environments? 2) \textbf{Robustness:} Does \method improve the base VLA’s robustness against OOD adversarial environments? 3) \textbf{Sim-to-real:} Does the improved robustness resulting from \method facilitate VLA sim-to-real transfer?

\subsection{Baselines}
\label{ssec:baselines}
Due to the lack of prior works that red-team VLAs through perturbing simulation environment parameters, we implement two environmental red-teaming baselines to ablate the effect of QD optimization employed by \method:
\begin{itemize}
    \item\textbf{Randomization:} This baseline samples environments from fixed Gaussian distributions centered at the default parameters of each LIBERO-Spatial task. The environments are rolled out to compute the benchmark metrics, but \textit{randomization} neither adapts its environment distributions nor performs QD search.
    \item\textbf{CMA-ES:} This baseline adapts the environment distributions using CMA-ES but optimizes only the failure rate. It therefore searches for adversarial environments without the diversity incentives provided by QD.
\end{itemize}
All methods search within the same 25-dimensional environment parameter space and use MIQP to repair invalid object placements (see Sec.~\ref{ssec:envgen}). The initial environment distribution is $\mathcal{N}(\theta_0, 0.02^2I)$, where $\theta_0$ is the default LIBERO-Spatial parameters. We use $k=4$ rollouts to compute the feedback for each environment and sample environments in batches of $B=8$. Since \method performs QD search using PCA-projected policy embeddings, we sample $32$ environments from $\mathcal{N}(\theta_0,0.04^2I)$ for each task and fit PCA to the corresponding policy embeddings. We then initialize $\mathcal{A}$ by discretizing the subspace spanned by the top $M=2$ principal components into $100\times100$ equally sized cells. We use $\alpha=0.1$ as \method's threshold learning rate.

\subsection{Simulation Experiments}
\label{ssec:simulation_exps}
We address the first two questions in simulation using LIBERO-Spatial~\cite{liu2023libero}, which consists of 10 language-conditioned pick-and-place tasks. We choose LIBERO-Spatial due to its prevalence as a VLA benchmark and its compatibility with our environment parameterizations.
To demonstrate \method{}’s application toward multiple VLA architectures, we experiment using two autoregressive models, $\pi_0$-FAST~\cite{pertsch2025fastefficientactiontokenization} and OpenVLA-OFT~\cite{kim2024openvla,kim2025fine}, and one flow-matching model,  $\pi_{0.5}$~\cite{intelligence2025pi05}. 

\noindent\textbf{Evaluating red-teaming:} We compare the methods using two metrics computed over the environments they generate under the same VLA evaluation budget ($3{,}200$ rollouts): (1) the \textit{number of feasibly adversarial environments} satisfying $0<\hat{P}^{\mathrm{fail}}<1$, and (2) the \textit{average pairwise DINOv2 embedding distance}. For the latter, we pass the first-step overview image from each environment through \texttt{dinov2\_vitb14} and compute the average pairwise cosine distance among the resulting embeddings. We use DINOv2 because its semantically structured representation space allows embedding distance to capture visual-semantic dissimilarity~\cite{oquab2023dinov2}.

\noindent\textbf{Evaluating robustness: } Following Sec.~\ref{ssec:fine_tuning}, we fine-tune the base VLA on successful trajectories collected from each method's rollouts. We train for $10{,}000$ steps using behavior cloning with rank-$16$ LoRA and a batch size of $32$. For each method, we reserve $20\%$ of its adversarial environments for evaluation and exclude their trajectories from the corresponding fine-tuning dataset. We evaluate the base and fine-tuned VLAs on the default LIBERO-Spatial tasks and the held-out environments from each method's evaluation set, and report the success rate for every VLA $\times$ evaluation set combination. Success rates on the $10$ LIBERO-Spatial tasks are evaluated over $50$ trials each, whereas success rates on held-out environments are evaluated over $4$ trials each due to the large number of environments.

\subsection{Real-World Experiments}
\label{ssec:real-world_exps}
To investigate whether \method improves VLA sim-to-real transfer, we construct a real-world test setting in which the VLA controls a Kinova JACO Gen2 arm to either \textit{grab the bowl from table center} (Task 1) or \textit{grab the bowl next to the plate} (Task 2). We modify LIBERO to create simulated digital twins of both tasks. To obtain a base VLA with reasonable success rates, we fine-tune the \texttt{pi05\_base} checkpoint on $40$ manually collected simulated demonstrations per task. We train for $10{,}000$ steps using behavior cloning with a batch size of $32$ and rank-$16$ LoRA. For each method, we then red-team the base VLA in simulation under a budget of $832$ VLA rollouts and fine-tune for another $10{,}000$ steps on successful trajectories to improve VLA robustness. The base model is also fine-tuned for another $10{,}000$ steps on the original demonstration to even the amount of training. We then evaluate the base and fine-tuned VLAs on the two tasks in the real world, and report success rates over $20$ trials on each VLA-task combination. A trial is successful if the robot navigates its gripper around the rim of the correct bowl within 3 minutes. We interleave trials between the two tested VLAs to control biases from external factors~\cite{kressgazit2024robotlearning}.

\section{Results}
\label{sec:results}

\begin{table*}[t]
\caption{VLA success rates on the default LIBERO-Spatial tasks and adversarial evaluation sets generated by randomization (rand), CMA-ES, and \method. Rows denote evaluation sets and columns denote VLA checkpoints.}
\label{tab:success_rates}
\centering
\resizebox{\linewidth}{!}{
\begin{tabular}{lcccccccccccc}
\toprule
 & \multicolumn{4}{c}{$\pi_{0}$\text{-FAST}} & \multicolumn{4}{c}{OpenVLA-OFT} & \multicolumn{4}{c}{$\pi_{0.5}$} \\
\cmidrule(lr){2-5} \cmidrule(lr){6-9} \cmidrule(lr){10-13}
Eval. set
 & base & rand & CMA-ES & Q-DEG & base & rand & CMA-ES & Q-DEG & base & rand & CMA-ES & Q-DEG \\
\midrule
default & \underline{96.80} & 68.60 & 82.60 & 89.60 & \underline{96.20} & 94.80 & 86.70 & 93.80 & \underline{98.40} & 85.60 & 93.60 & 95.60 \\
rand & 29.15 & 39.40 & 37.79 & \underline{48.73} & 56.25 & \underline{82.29} & 68.75 & 78.13 & 77.38 & 77.98 & 83.33 & \underline{88.09} \\
CMA-ES        & 27.90 & \underline{38.46} & 24.81 & 29.82 & 45.75 & 52.52 & 48.27 & \underline{75.79} & 54.98 & 49.39 & \underline{69.17} & 58.00 \\
Q-DEG       & 54.83 & 51.28 & 36.89 & \underline{56.13} & 47.19 & 50.77 & 55.61 & \underline{69.13} & 38.03 & 61.82 & 70.45 & \underline{81.81} \\
\midrule
average       & 52.17 & 49.44 & 45.52 & \textcolor{green!50!black}{\textbf{56.07}} & 61.35 & 70.10 & 64.83 & \textcolor{green!50!black}{\textbf{79.21}} & 67.20 & 68.70 & 79.14 & \textcolor{green!50!black}{\textbf{80.88}}  \\
\bottomrule
\end{tabular}
}
\vspace{-5pt}
\end{table*}

\begin{figure}[h]
  \centering
  \includegraphics[width=\linewidth]{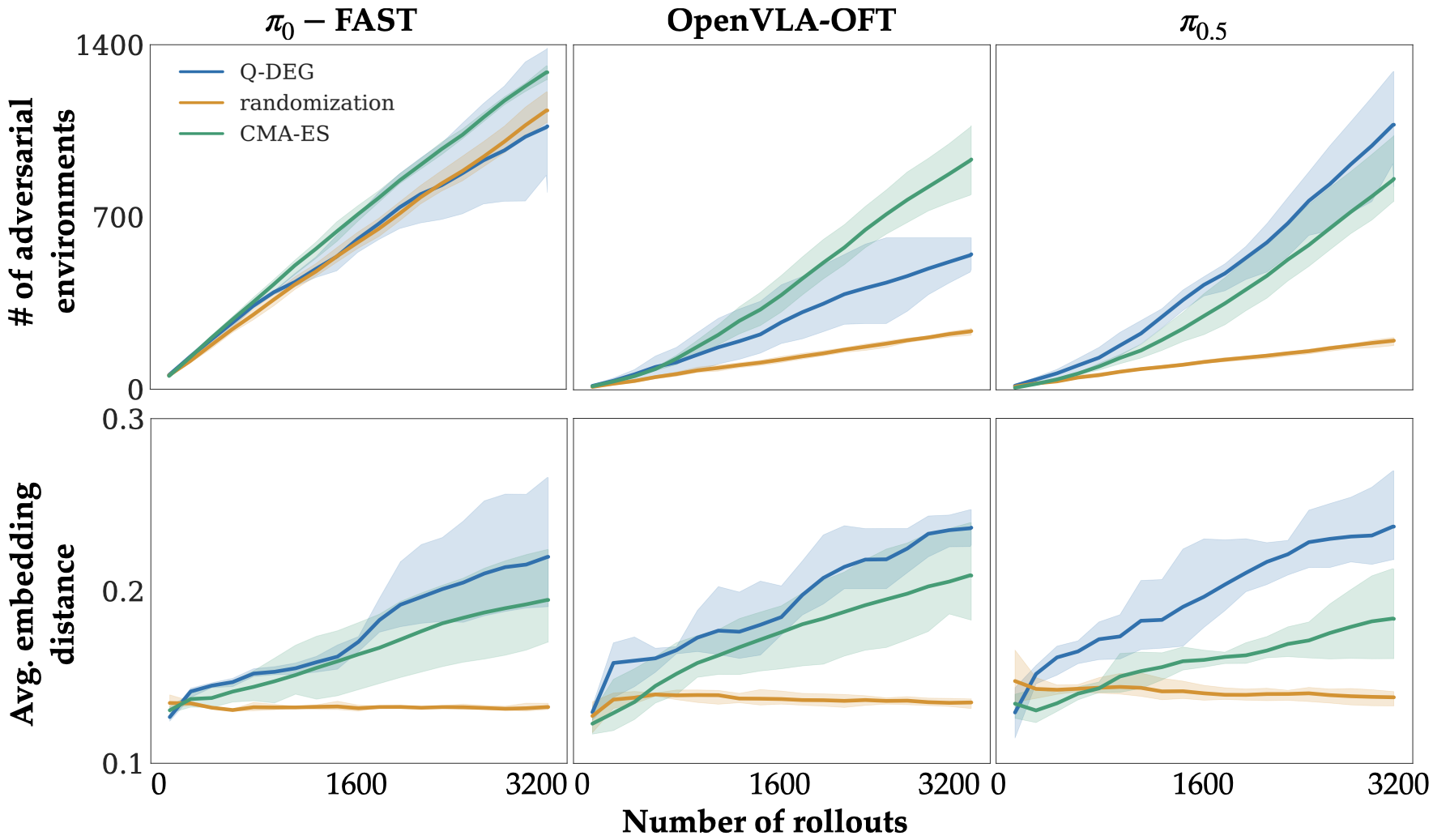}
  \caption{Number of feasibly adversarial environments (top) found by \textit{randomization}, \textit{CMA-ES}, and \method, and the corresponding average pairwise DINOv2 cosine distance among environments (bottom), plotted against the number of VLA rollouts. Shaded regions denote $95\%$ confidence intervals aggregated over three independent trials.}
  \label{fig:scores}
  \vspace{-10pt}
\end{figure}

\noindent\textbf{Red-teaming:}
As shown in Figure~\ref{fig:scores}, all methods find hundreds of feasibly adversarial environments in which the VLAs fail, revealing their vulnerabilities to our environmental perturbations despite their near-perfect performance on default LIBERO-Spatial. \method consistently finds environments more diverse than the baselines across VLAs, as reflected by higher pairwise embedding distances. \textit{Randomization} does not adapt its environment distribution, so it only generates near-distribution environments around the default LIBERO-Spatial configurations. Such environments not only lack diversity but also induce few failures in the stronger OpenVLA-OFT and $\pi_{0.5}$ VLAs. On the other hand, the failure-maximizing baseline \textit{CMA-ES} matches \method in the number of feasibly adversarial environments found, but its environments tend to cluster around a few configurations that induce the most VLA failures, resulting in lower environmental diversity. The advantage of \method over baselines is most evident on $\pi_{0.5}$, for which finding feasibly adversarial environments is more challenging due to the model's robustness. In this case, \method's environmental diversity translates into a larger number of environments found, outperforming even the failure-maximizing baseline \textit{CMA-ES}. This can be explained by the role of diverse environments as ``stepping stones'' that enable exploration toward niche configurations. Figure~\ref{fig:pull} shows examples of adversarial environments found by \method on $\pi_{0.5}$. Additional visualizations are available on our anonymous project website: \url{https://qd-env.github.io/qdeg/}.

\noindent\textbf{Robustness:} 
Table~\ref{tab:success_rates} reports VLA success rates across evaluation sets and checkpoints. Rows denote the default LIBERO-Spatial tasks and the held-out adversarial environments generated by \textit{randomization}, \textit{CMA-ES}, and \method, while columns denote the base VLA and the VLAs fine-tuned on successful trajectories collected from each method's rollouts. Across all three VLA models, \method consistently improves the base VLA in adversarial evaluation sets, with gains up to $43.78\%$. Compared to VLAs fine-tuned using baseline methods, VLAs fine-tuned using \method also generalize better between evaluation sets, often matching or outperforming baselines even on their respective evaluation sets. These results suggest that the lack of diversity in the fine-tuning datasets collected by baselines causes their VLAs to overcommit to a narrower range of failure modes. Fine-tuning on additional adversarial environments causes some forgetting, so all methods slightly reduce success rates on the default LIBERO-Spatial tasks. However, \method causes less degradation than baselines, and this minor drop is outweighed by substantial robustness gains on adversarial evaluation sets.

\begin{figure}[h]
  \centering
  \includegraphics[width=0.99\linewidth]{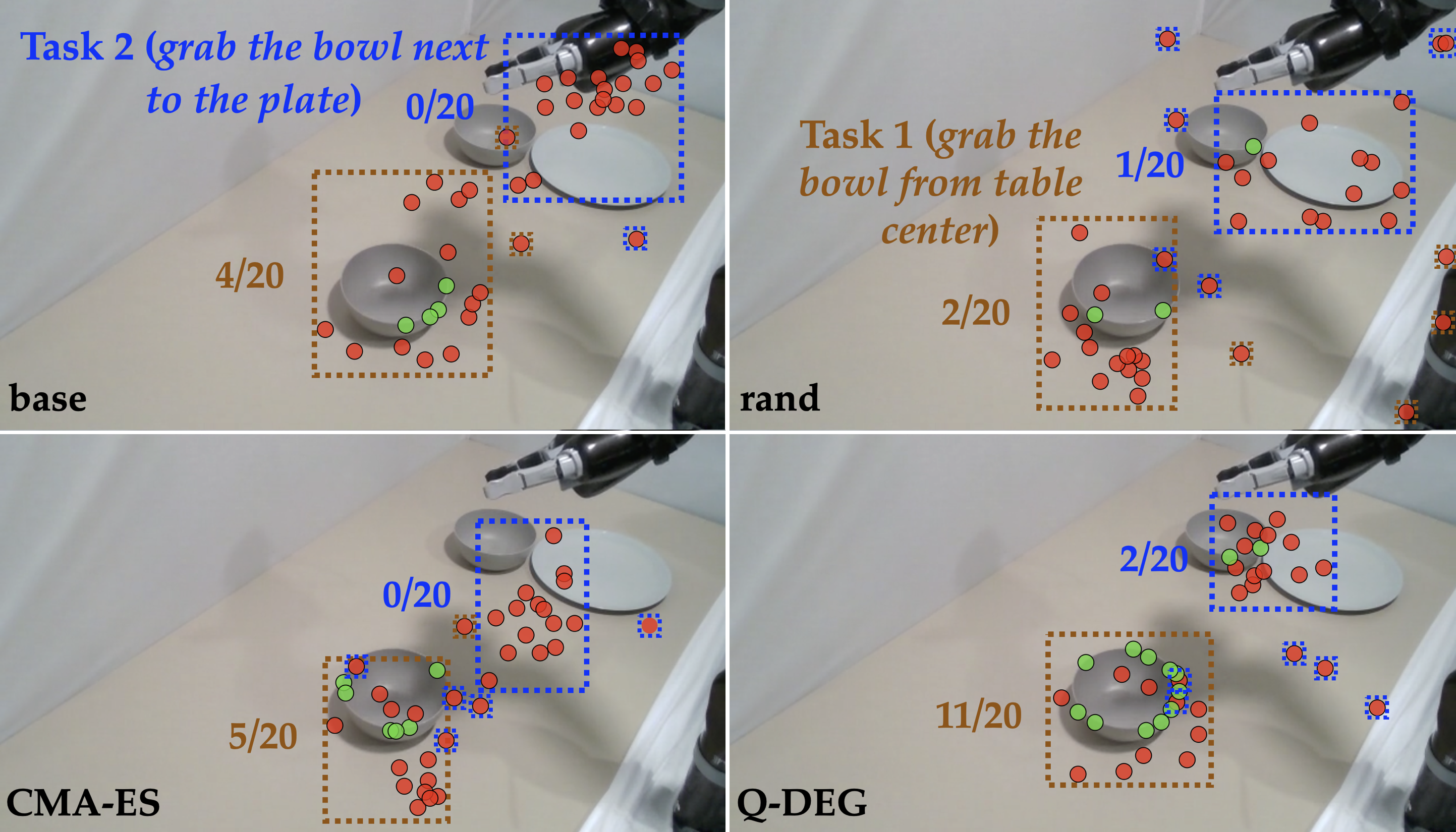}
  \caption{Success counts and final gripper positions when evaluating the base VLA and VLAs fine-tuned with \textit{randomization} (\textit{rand}), \textit{CMA-ES}, and \method on real-world tasks. Successful trials are shown in green, and failed trials are shown in red.}
  \label{fig:real}
  \vspace{-10pt}
\end{figure}

\noindent\textbf{Sim-to-real:} 
Figure~\ref{fig:real} reports real-world success counts for the base VLA trained on the JACO tasks in simulation and the VLAs further fine-tuned for robustness using trajectories from \textit{Randomization}, \textit{CMA-ES}, and \method. The final gripper position in each trial is marked with a green dot for success or a red dot for failure. The gripper positions of the base VLA and the VLAs fine-tuned using the baseline methods often cluster at incorrect locations. For example, in Task 2, the base VLA frequently moves toward the empty region to the upper right of the target bowl, whereas the VLA fine-tuned using \textit{CMA-ES} often ends up to its lower right. These clusters suggest that the VLAs overfit to spatial patterns in their respective fine-tuning datasets, weakening sim-to-real transfer. In contrast, fine-tuning with \method improves robustness to variations in object placement, allowing the VLA to navigate the gripper more consistently toward the target bowl rather than toward fixed locations.

\vspace{-5pt}

\section{Conclusion and Limitations}
\label{sec:conclusion}
In this work, we present \method, a method that red-teams VLAs by using Quality Diversity optimization to search for diverse, feasibly adversarial environments, and improves robustness by fine-tuning the VLA on its own successful rollouts in those environments.
Across three representative VLA models, \method finds more diverse adversarial environments than baselines and improves VLA robustness to OOD environments in both simulated and real-world manipulation tasks, outperforming the base VLA and VLAs fine-tuned on baseline data.

A limitation with \method is that it fine-tunes only on successful rollouts, and does not use failed trajectories despite their information about actions to avoid. %
Future work could incorporate these failures by forming pairs of successful and failed trajectories for preference learning methods~\cite{rafailov2023direct}. 
A second limitation is that \method uses PCA to reduce the dimensionality of policy embeddings before the discretization-based CMA-MAE search, potentially discarding task-relevant information. Recent discretization-free QD algorithms, such as Soft QD~\cite{hedayatian2025soft}, could obviate this dimensionality reduction and enable diversification in full embedding space.

\appendices
\section{Policy Embedding Extraction and Compression}
\label{app:polemb}
We extract embeddings from $\pi_0$-FAST, OpenVLA-OFT, and $\pi_{0.5}$ as follows:

\noindent\textbf{$\pi_0$-FAST:} During autoregressive action decoding, the final transformer block preceding the action head in $\pi_0$-FAST produces a $2048$-dimensional hidden state at each FAST-token decoding step. We average the hidden states across all decoding steps prior to the EOS token, yielding a $2048$-dimensional policy embedding.

\noindent\textbf{OpenVLA-OFT:} The final transformer block preceding the action head outputs hidden states of shape $\mathbb{R}^{d_a \times d_c \times 4096}$, where $d_a$ and $d_c$ denote the action dimension and action chunk size. We average over the first two dimensions to obtain a $4096$-dimensional policy embedding.

\noindent$\boldsymbol{\pi_{0.5}:}$ Unlike the $\pi_0$-FAST and OpenVLA-OFT, $\pi_{0.5}$ incorporates robot state information as part of its action expert during conditional flow-matching. To ensure that the policy embeddings capture all information relevant to action generation, we extract activations from the final layer before velocity prediction at the first Euler integration step, yielding a $1024$-dimensional policy embedding. Because flow-matching is initialized from a standard Gaussian source distribution, the resulting activations can exhibit stochastic variation unrelated to the observation. To obtain more consistent embeddings, we perform a separate flow evaluation initialized from the zero vector rather than a Gaussian sample. This auxiliary evaluation is used solely for embedding extraction and does not affect the actions generated by the policy.

\begin{figure}[h]
  \centering
  \includegraphics[width=\linewidth]{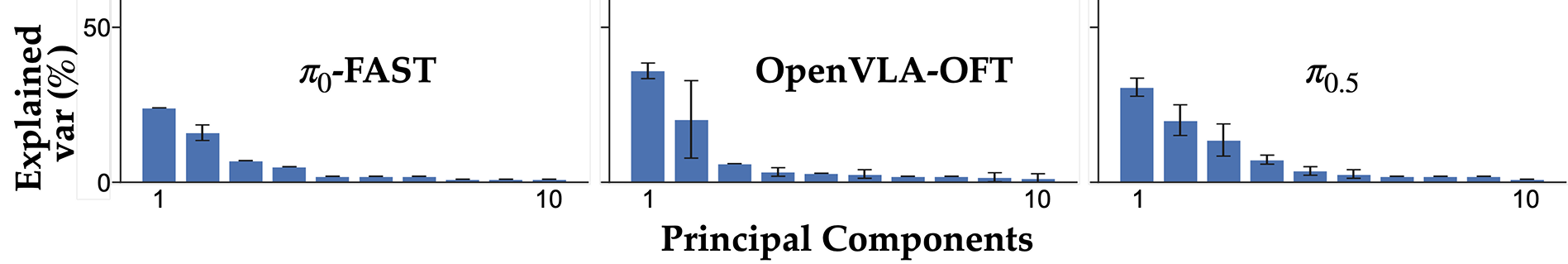}
  \caption{Variance explained by the first 10 principal components of raw policy embeddings extracted from $\pi_0$-FAST, OpenVLA-OFT, and $\pi_{0.5}$. Results are aggregated over three independent trials, each sampling $320$ environments. Error bars denote $95\%$ confidence intervals.}
  \label{fig:pc}
\end{figure}

Discretizing the high-dimensional raw policy embedding space would require a prohibitively large number of QD archive cells to achieve adequate coverage~\cite{paolo2020unsupervised, hedayatian2025soft}. Following prior work~\cite{grillotti2022unsupervised}, we therefore fit PCA to the raw policy embeddings and project them onto the low-dimensional subspace spanned by the top \(M=2\) principal components. Although this projection discards substantial information, we find that the first two principal components consistently capture a large fraction of the variation among policy embeddings collected from sampled environments $\mathcal{E}_\mathbf{c}(\theta \sim \mathcal{N}(\theta_0,0.04^2I))$, explaining approximately $50\%$ of the total variance, while subsequent components contribute substantially less, as shown in Figure~\ref{fig:pc}.

\section{Uniform Sampling Baseline}
\label{app:uniform}
We explored an additional baseline \emph{uniform}, which was similar to \emph{rand}, but sampled parameters uniformly within each dimension's validity range instead of near the default values. \emph{Uniform} generated diverse but OOD environments on which VLAs failed consistently, so few of \emph{uniform}'s environments counted as feasibly adversarial. After $3{,}200$ rollouts, \emph{uniform} yielded $0 \pm 0$, $29.0 \pm 7.9$, and $179.3 \pm 13.5$ feasibly adversarial environments for $\pi_0$-FAST, OpenVLA-OFT, and $\pi_{0.5}$, respectively, across three trials. The corresponding average pairwise DINOv2 embedding distances were $0 \pm 0$, $0.255 \pm 0.015$, and $0.291 \pm 0.002$. After robustness fine-tuning on successful trajectories, \emph{uniform}'s $\pi_{0.5}$ checkpoint achieved success rates of $94.00\%$, $76.79\%$, $60.19\%$, and $71.67\%$ on the default, \emph{rand}, \emph{CMA-ES}, and \method evaluation sets, respectively. We did not run robustness fine-tuning experiments on $\pi_0$-FAST or OpenVLA-OFT due to \emph{uniform}'s low numbers of feasibly adversarial environments, and by extension self-imitable trajectories on these models. Overall, \emph{uniform} performed poorly in both red-teaming and robustness improvement for the weaker VLA models and was only comparable to \emph{rand} on $\pi_{0.5}$.

\bibliographystyle{IEEEtran}
\bibliography{references}

\end{document}